\documentclass{article}

\usepackage{graphicx} 
\usepackage{bold-extra}
\usepackage{subfigure} 
\usepackage{amsmath} 
\usepackage{amsfonts}

\newcommand{\qed}{\nobreak \ifvmode \relax \else
      \ifdim\lastskip<1.5em \hskip-\lastskip
      \hskip1.5em plus0em minus0.5em \fi \nobreak
      \vrule height0.75em width0.5em depth0.25em\fi}

\usepackage{natbib}

\usepackage{algorithm}
\usepackage{algorithmic}

\usepackage{hyperref}

\usepackage[accepted]{iclr2014} 

\icmltitlerunning{Catastrophic Forgetting}

\begin{document} 

\twocolumn[
\icmltitle{An Empirical Investigation of Catastrophic Forgetting in Gradient-Based Neural Networks}
\icmlauthor{Ian J. Goodfellow}{goodfeli@iro.umontreal.ca}
\icmlauthor{Mehdi Mirza}{mirzamom@iro.umontreal.ca}
\icmlauthor{Da Xiao}{xiaoda99@bupt.edu.cn}
\icmlauthor{Aaron Courville}{aaron.courville@umontreal.ca}
\icmlauthor{Yoshua Bengio}{yoshua.bengio@umontreal.ca}


\icmlkeywords{multilayer perceptron, convolutional network, classification, computer vision}

\vskip 0.3in
]

\begin{abstract}
{\em Catastrophic forgetting} is a problem faced by many machine learning models and algorithms.
When trained on one task, then trained on a second task, many machine learning models ``forget''
how to perform the first task. This is widely believed to be a serious problem for neural networks.
Here, we investigate the extent to which the catastrophic forgetting problem occurs for modern neural
networks, comparing both established and recent gradient-based training algorithms and activation
functions. We also examine the effect of the relationship between the first task and the second task
on catastrophic forgetting. We find that it is always best to train using the dropout algorithm--the
dropout algorithm is consistently best at adapting to the new task, remembering the old task, and
has the best tradeoff curve between these two extremes. We find that different tasks and relationships
between tasks result in very different rankings of activation function performance. This suggests that
the choice of activation function should always be cross-validated.
\end{abstract} 

\section{Introduction}

Catastrophic forgetting\citep{McCloskeyCohen89,Ratcliff90} is a problem that affects neural networks,
as well as other learning systems, including both biological and machine learning systems. When a learning
system is first trained on one task, then trained on a second task, it may forget how to perform the
first task. For example, a machine learning system trained with a convex objective will always reach the
same configuration at the end of training on the second task, regardless of how it was initialized. This
means that an SVM that is trained on two different tasks will completely forget how to perform the first
task. Whenever the SVM is able to correctly classify an example from the original task, it is only due to
chance similarities between the two tasks.

A well-supported model of biological learning in human beings suggests that neocortical neurons learn using
an algorithm that is prone to catastrophic forgetting, and that the neocortical learning algorithm is complemented
by a virtual experience system that replays memories stored in the hippocampus in order to continually reinforce
tasks that have not been recently performed~\citep{McClellandMcNaughtonOReilly95}. As machine learning researchers,
the lesson we can glean from this is that it is acceptable for our learning algorithms to suffer from forgetting, but
they may need complementary algorithms to reduce the information loss. Designing such complementary algorithms
depends on understanding the characteristics of the forgetting experienced by our contemporary primary learning
algorithms.

In this paper we investigate the extent to which catastrophic forgetting affects a variety
of learning algorithms and neural network activation functions. Neuroscientific evidence
suggests that the relationship between the old and new task strongly influences the outcome
of the two successive learning experiences~\citep{McClelland13}. Consequently, we examine three
different types of relationship between tasks: one in which the tasks are functionally identical
but with different formats of the input, one in which the tasks are similar, and one in which the
tasks are dissimilar.

We find that dropout~\citep{Hinton-et-al-arxiv2012} is consistently the best training algorithm for
modern feedforward neural nets.
The choice of activation function has a less consistent effect--different activation functions are
preferable depending on the task and relationship between tasks, as well as whether one places greater
emphasis on adapting to the new task or retaining performance on the old task. When training with dropout,
maxout~\citep{Goodfellow-et-al-ICML2013} is the only activation function to consistently appear
somewhere on the frontier of performance tradeoffs for all tasks we considered. However,
maxout is not the best function at all points along
the tradeoff curve, and does not have as consistent performance when trained without dropout, so it is
still advisable to cross-validate the choice of activation function, particularly when training without
dropout.

We find that in most cases, dropout increases the optimal size of the net, so the resistance to forgetting
may be explained mostly by the larger nets having greater capacity. However, this effect is not consistent,
and when using dissimilar task pairs, dropout usually decreases the size of the net. This suggests dropout
may have other more subtle beneficial effects to characterize in the future.

\section{Related work}

Catastrophic forgetting has not been a well-studied property of neural networks in recent years. This
property was well-studied in the past, but has not received much attention since the deep learning
renaissance that began in 2006. \citet{Srivastava13} re-popularized the idea of studying this aspect
of modern deep neural nets.

However, the main focus of this work was not to study catastrophic forgetting, so the experiments
were limited. Only one neural network was trained in each case. The networks all used the same
hyperparameters, and the same heuristically chosen stopping point. Only one pair of tasks
was employed, so it is not clear whether the findings apply only to pairs of tasks with the same kind
and degree of similarity or whether the findings generalize to many kinds of pairs of tasks.
Only one training algorithm, standard gradient descent was employed. We move beyond all of these
limitations by training multiple nets with different hyperparameters, stopping using a validation set,
evaluating using three task pairs with different task similarity profiles, and including the dropout algorithm
in our set of experiments.

\section{Methods}

In this section, we describe the basic algorithms and techniques used in our experiments.

\subsection{Dropout}

Dropout~\citep{Hinton-et-al-arxiv2012,Srivastava-master-small} is a recently introduced training
algorithm for neural networks. Dropout is designed to regularize neural networks in order to improve
their generalization performance.

Dropout training is a modification to standard stochastic gradient descent training. When each example
is presented to the network during learning, the input states and hidden unit states of the network
are multiplied by a binary mask. The zeros in the mask cause some units to be removed from the network.
This mask is generated randomly each time an example is presented. Each element of the mask is sampled
independently of the others, using some fixed probability $p$. At test time, no units are dropped, and
the weights going out of each unit are multiplied by $p$ to compensate for that unit being present more
often than it was during training.

Dropout can be seen as an extremely efficient means of training exponentially many neural networks that
share weights, then averaging together their predictions. This procedure resembles bagging, which helps
to reduce the generalization error. The fact that the learned features must work well in the context of
many different models also helps to regularize the model.

Dropout is a very effective regularizer. Prior to the introduction of dropout, one of the main ways
of reducing the generalization error of a neural network was simply to restrict its capacity by using a
small number of hidden units. Dropout enables training of noticeably larger networks. As an example, we
performed random hyperparameter search with 25 experiments in each case to find the best two-layer rectifier
network~\citep{Glorot+al-AI-2011} for classifying the MNIST dataset. When training with dropout, the best
network according to the validation set had 56.48\% more parameters than the best network trained without
dropout.

We hypothesize that the increased size of optimally functioning dropout nets means that they are less
prone to the catastrophic forgetting problem than traditional neural nets, which were regularized by
constraining the capacity to be just barely sufficient to perform the first task. 

\subsection{Activation functions}

Each of the hidden layers of our neural networks transforms some input vector $x$ into an output vector $h$.
In all cases, this is done by first computing a {\em presynaptic activation} $z = W x + b$
where $W$ is a matrix of learnable parameters and $b$ is a vector of learnable parameters.
The presynaptic activation $z$ is then transformed into a post-synaptic activation $h$ by
an {\em activation function}: $h = f(z)$. $h$ is then provided as the input to the next layer.

We studied the following activation functions:

\begin{enumerate}
\item {\em Logistic sigmoid}: 
\[ \forall i, f(z)_i = \frac { 1 } { 1 + \exp(-z_i) } \]

\item {\em Rectified linear}~\citep{Jarrett-ICCV2009,Glorot+al-AI-2011}:
\[ \forall i, f(z)_i = \text{max}(0, z_i) \]

\item {\em Hard Local Winner Take All (LWTA)} \citep{Srivastava13}:
\[ \forall i, f(z)_i = g(i, z) z_i .\]
Here $g$ is a gating function. $z$ is divided into disjoint blocks of size $k$,
and $g(i, z)$ is 1 if $z_i$ is the maximal element of its group. If more than
one element is tied for the maximum, we break the tie uniformly at random%
~\footnote{This is a deviation from the implementation of \citet{Srivastava13},
who break ties by using the smallest index. We used this approach because it is
easier to implement in Theano. We think our deviation from the previous implementation
is acceptable because we are able to reproduce the previously reported classification
performance.
}%
. Otherwise $g(i, z)$ is 0. 

\item {\em Maxout}~\citep{Goodfellow-et-al-ICML2013}:
\[ \forall i, f(z)_i = \text{max}_j \{ z_{ki}, \dots, z_{k(i+1) -1} \} \]

\end{enumerate}

We trained each of these four activation functions with each of the two algorithms we considered, for
a total of eight distinct methods.

\subsection{Random hyperparameter search}

Making fair comparisons between different deep learning methods is difficult. The performance of most
deep learning methods is a complicated non-linear function of multiple hyperparameters. For many applications,
the state of the art performance is obtained by a human practitioner selecting hyperparameters for some
deep learning method. Human selection is problematic for comparing methods because the human practitioner
may be more skillful at selecting hyperparameters for methods that he or she is familiar with. Human practitioners
may also have a conflict of interest predisposing them to selecting better hyperparameters for methods that they
prefer.

Automated selection of hyperparameters allows more fair comparison of methods with a complicated dependence on
hyperparameters. However, automated selection of hyperparameters is challenging. Grid search suffers from the
curse of dimensionality, requiring exponentially many experiments to explore high-dimensional hyperparameter
spaces. In this work, we use random hyperparameter search~\citep{Bergstra+Bengio-2012} instead. This method is
simple to implement and obtains roughly state of the art results using only 25 experiments on simple datasets
such as MNIST.

Other more sophisticated methods of hyperparameter search, such as Bayesian optimization, may be able to obtain
better results, but we found that random search was able to obtain state of the art performance on the tasks we
consider, so we did not think that the greater complication of using these methods was justified. More sophisticated
methods of hyperparameter feedback may also introduce some sort of bias into the experiment, if one of the methods
we study satisfies more of the modeling assumptions of the hyperparameter selector.

\section{Experiments}

All of our experiments follow the same basic form. For each experiment, we define two tasks: the ``old task'' and
the ``new task.'' We examine the behavior of neural networks that are trained on the old task, then trained on the
new task.

For each definition of the tasks, we run the same suite of experiments for two kinds of algorithms: stochastic gradient
descent training, and dropout training. For each of these algorithms, we try four different activation functions: logistic
sigmoid, rectifier, hard LWTA, and maxout.

For each of these eight conditions, we randomly generate 25 random sets of hyperparameters. See the code accompanying
the paper for details. In all cases, we use a model with two hidden layers followed by a softmax classification layer. The
hyperparameters we search over include the magnitude of the max-norm constraint~\citep{Srebro05} for each layer, the method
used to initialize the weights for each layer and any hyper-parameters associated with such method, the initial biases for
each layer, the parameters controlling a saturating linear learning rate decay and momentum increase schedule, and the size
of each layer.

We did not search over some hyperparameters for which good values are reasonably well-known. For example, for dropout, the
best probability of dropping a hidden unit is known to usually be around 0.5, and the best probability of dropping a visible
unit is known to usually be around 0.2. We used these well-known constants on all experiments. This may reduce the maximum
possible performance we are able to obtain using our search, but it makes the search function much better with only 25 experiments
since fewer of the experiments fail dramatically.

We did our best to keep the hyperparameter searches comparable between different methods. We always used the same hyperparameter
search for SGD as for dropout. For the different activation functions, there are some slight differences between the hyperameter
searches. All of these differences are related to parameter initialization schemes. For LWTA and maxout, we always set the initial
biases to 0, since randomly initializing a bias for each unit can make one unit within a group win the max too often, resulting in
dead filters. For rectifiers and sigmoids, we randomly select the initial biases, but using different distributions. Sigmoid networks
can benefit from significantly negative initial biases, since this encourages sparsity, but these initializations are fatal to rectifier
networks, since a significantly negative initial bias can prevent a unit's parameters from ever receiving non-zero gradient. Rectifier
units can also benefit from slightly positive initial biases, because they help prevent rectifier units from getting stuck, but there
is no known reason to believe this helps sigmoid units. We thus use a different range of initial biases for the rectifiers and the sigmoids.
This was necessary to make sure that each method is able to achieve roughly state of the art performance with only 25 experiments in
the random search. Likewise, there are some differences in the way we initialize the weights for each activation function. For all activation
functions, we initialize the weights from a uniform distribution over small values, in at least some cases. For maxout and LWTA, this is always
the method we use. For rectifiers and sigmoids, the hyperparameter search may also choose to use the initialization method advocated
by \citet{Martens+Sutskever-ICML2011}. In this method, all but $k$ of the weights going into a unit are set to 0, while the remaining $k$ are
set to relatively large random values. For maxout and LWTA, this method performs poorly because different filters within the same group can
be initialized to have extremely dissimilar semantics.

In all cases, we first train on the ``old task'' until the validation set error has not improved in the last 100 epochs. Then we restore
the parameters corresponding to the best validation set error, and begin training on the ``new task''. We train until the error on the union
of the old validation set and new validation set has not improved for 100 epochs.

After running all 25 randomly configured experiments for all 8 conditions, we make a possibilities frontier curve showing the
minimum amount of test error on the new task obtaining for each amount of test error on the old task. Specifically, these plots
are made by drawing a curve that traces out the lower left frontier of the cloud of points of all (old task test error, new task test error)
pairs encountered by all 25 models during the course of training on the new task, with one point generated after each pass through the
training set.
Note that these test set errors are
computed after training on only a subset of the training data, because we do not train on the validation set. It is possible to improve
further by also training on the validation set, but we do not do so here because we only care about the relative performance of the
different methods, not necessarily obtaining state of the art results.

(Usually possibilities frontier curves are used in scenarios where higher values are better, and the curves trace out the higher edge of a convex hull of scatterplot. Here,
we are plotting error rates, so the lower values are better and the curves trace out the lower edge of a convex hull of a scatterplot. We used error
rather than accuracy so that log scale plots would compress regions of bad performance and expand regions of good performance, in order
to highlight the differences between the best-performing methods. Note that the log scaling sometimes makes the convex regions apear non-convex)

\subsection{Input reformatting}

Many naturally occurring tasks are highly similar to each other in terms of the underlying structure that must be understood, but
have the input presented in a different format.

For example, consider learning to understand Italian after
already learning to understand Spanish. Both tasks share the deeper underlying structure of being a natural
language understanding problem, and furthermore, Italian and Spanish have similar grammar. However, the specific
words in each language are different. A person learning Italian thus benefits from having a pre-existing representation
of the general structure of the language. The challenge is to learn to map the new words into these structures
(e.g., to attach the Italian word ``sei'' to the pre-existing concept of the second person conjugation of the verb ``to be'') without
damaging the ability to understand Spanish. The ability to understand Spanish could diminish if the learning algorithm
inadvertently modifies the more abstract definition of language in general (i.e., if neurons that were used for verb
conjugation before now get re-purposed for plurality agreement) rather than exploiting the pre-existing definition,
or if the learning algorithm removes the associations between individual Spanish words and these pre-existing
concepts (e.g., if the net retains the concept of there being a second person conjugation of the verb ``to be'' but forgets
that the Spanish word ``eres'' corresponds to it).

To test this kind of learning problem, we designed a simple pair of tasks, where the tasks are the same, but with different ways
of formatting the input.
Specifically, we used MNIST classification, but with a different permutation of the pixels for the old task and the new task.
Both tasks thus benefit from having concepts like penstroke detectors, or the concept of penstrokes being combined to form digits.
However, the meaning of any individual pixel is different. The net must learn to associate new collections of pixels to penstrokes,
without significantly disrupting the old higher level concepts, or erasing the old connections between pixels and penstrokes.

\begin{figure*}
\includegraphics[width=\textwidth]{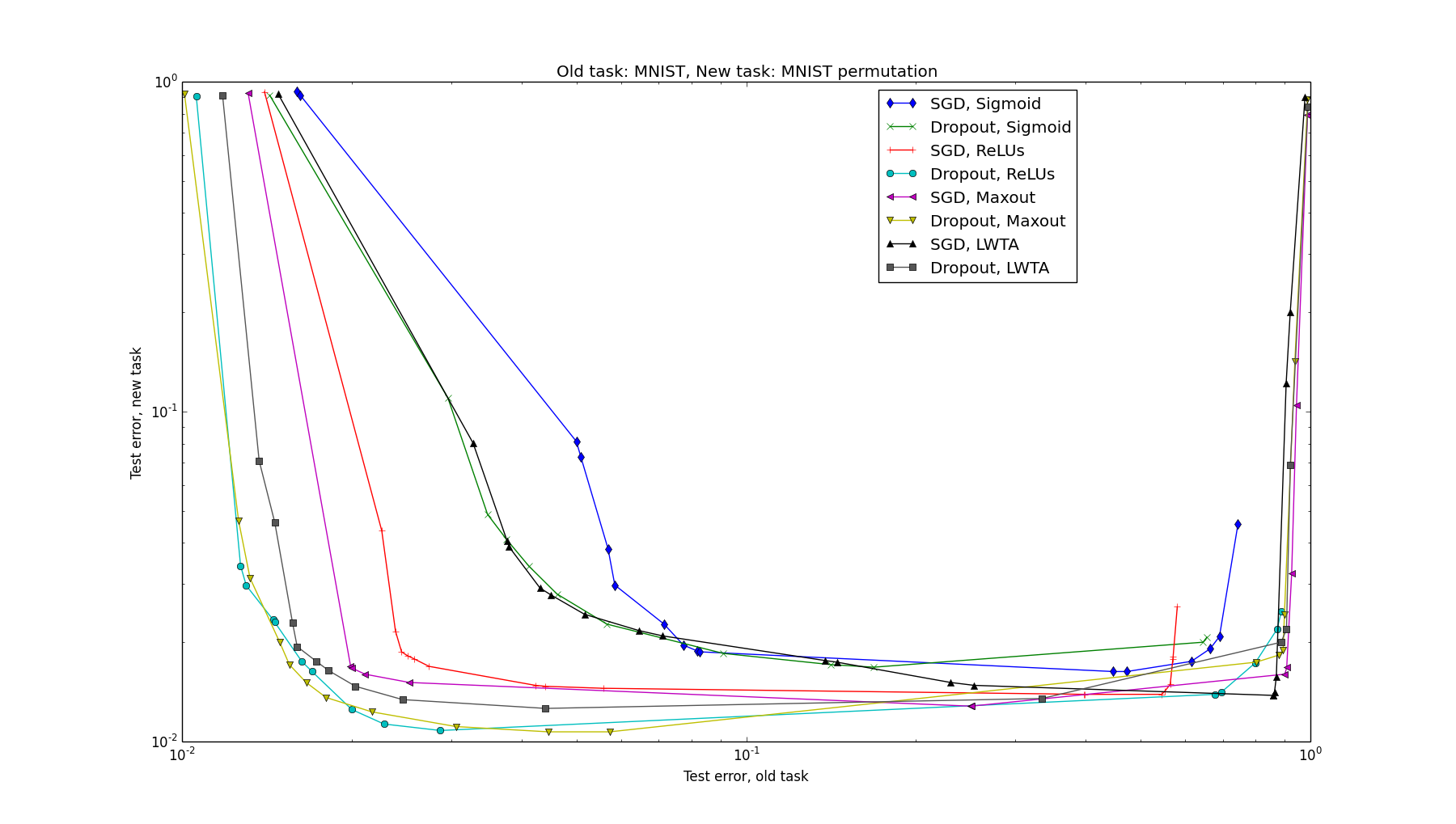}
\caption{Possibilities frontiers for the input reformatting experiment.}
\label{input_reformatting}
\end{figure*}

\begin{figure}
\centering
\includegraphics[width=.4\textwidth]{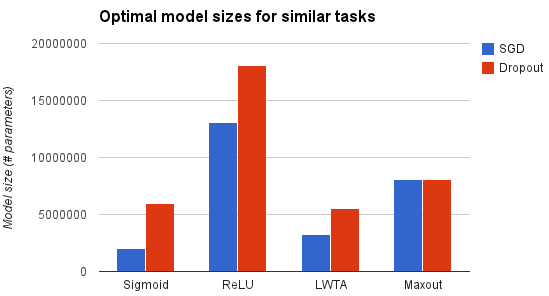}
\caption{Optimal model size with and without dropout on the input reformatting tasks.}
\label{input_reformatting_size}
\end{figure}

The classification performance results are presented
in Fig.~\ref{input_reformatting}.
Using dropout improved the two-task validation set performance for all models on this task pair.
We show the effect of dropout on the optimal model size in Fig.~\ref{input_reformatting_size}.
While the nets were able to basically succeed at this task, we don't believe that they did so
by mapping different sets of pixels into pre-existing concepts. We visualized the first layer weights
of the best net (in terms of combined validation set error) and their apparent semantics do not noticeably
change between when training on the old task concludes and training on the new task begins. This suggests
that the higher layers of the net changed to be able to accomodate a relatively arbitrary projection of the
input, rather than remaining the same while the lower layers adapted to the new input format.

\subsection{Similar tasks}

We next considered what happens when the two tasks are not exactly the same, but semantically similar, and using the same input format. To test this case,
we used sentiment analysis of two product categories of Amazon reviews~\citep{Blitzer2007} as the two tasks. The task is just to classify the text of a
product review as positive or negative in sentiment. We used the same preprocessing as~\citep{Glorot+al-ICML-2011}.

\begin{figure*}
\includegraphics[width=\textwidth]{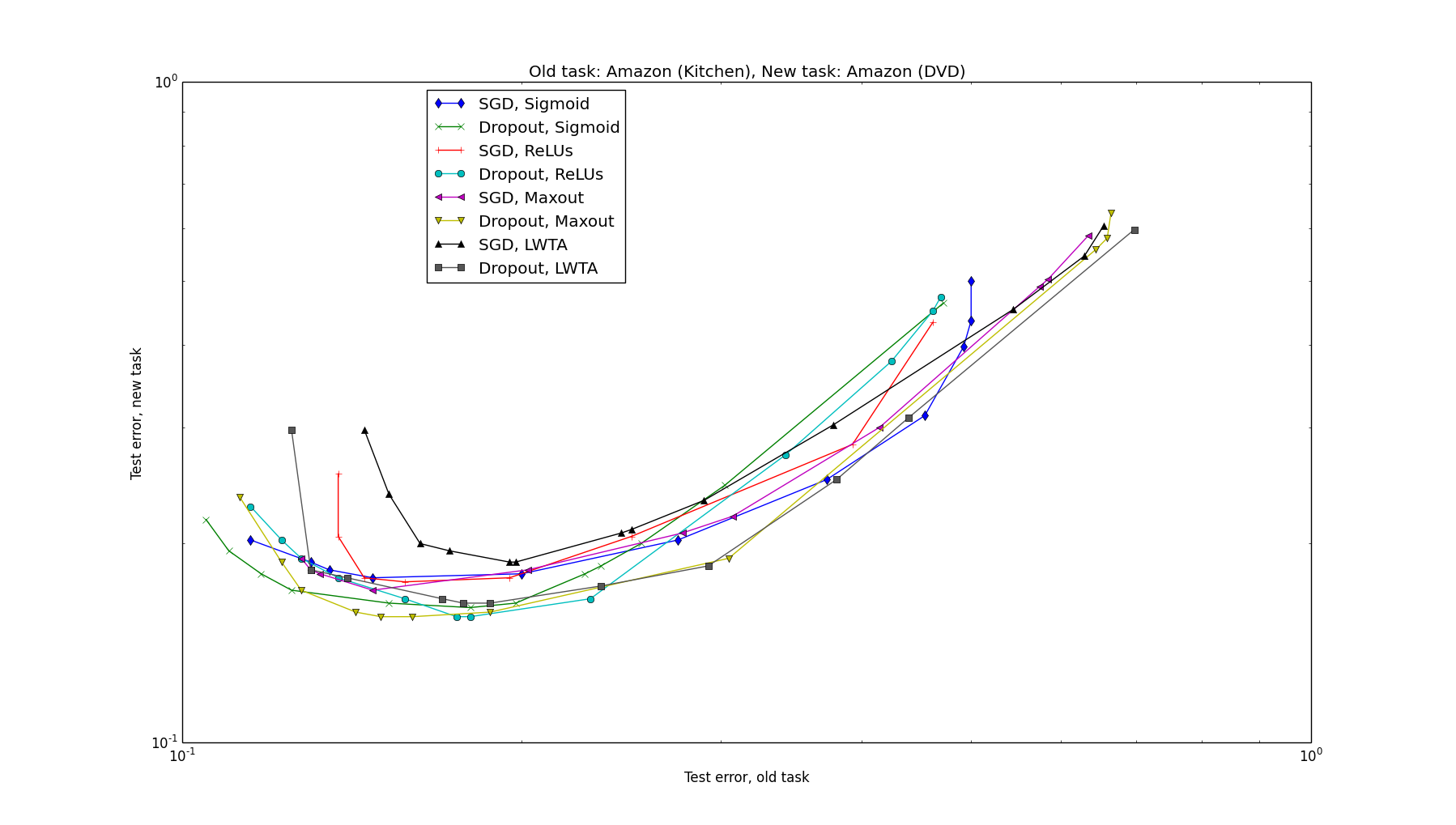}
\caption{Possibilities frontiers for the similar tasks experiment.}
\label{similar_tasks}
\end{figure*}

The classification performance results are presented
in Fig.~\ref{similar_tasks}.
Using dropout improved the two-task validation set performance for all models on this task pair.
We show the effect of dropout on the optimal model size in Fig.~\ref{similar_size}.

\begin{figure}
\centering
\includegraphics[width=.4\textwidth]{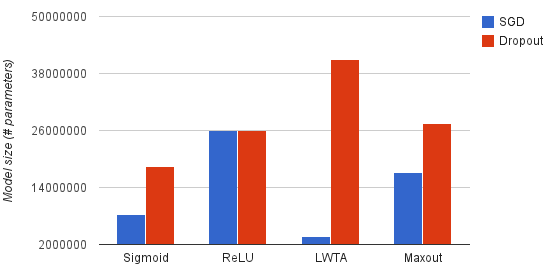}
\caption{Optimal model size with and without dropout on the similar tasks experiment.}
\label{similar_size}
\end{figure}

\subsection{Dissimilar tasks}

We next considered what happens when the two tasks are semantically similar. To test this case,
we used Amazon reviews as one task, and MNIST classification as another. In order to give both
tasks the same output size, we used only two classes of the MNIST dataset. To give them the
same validation set size, we randomly subsampled the remaining examples of the MNIST validation
set (since the MNIST validation set was originally larger than the Amazon validation set, and we
don't want the estimate of the performance on the Amazon dataset to have higher variance than the
MNIST one). The Amazon dataset as we preprocessed it earlier has 5,000 input features, while MNIST
has only 784. To give the two tasks the same input size, we reduced the dimensionality of the Amazon
data with PCA.

\begin{figure*}
\includegraphics[width=\textwidth]{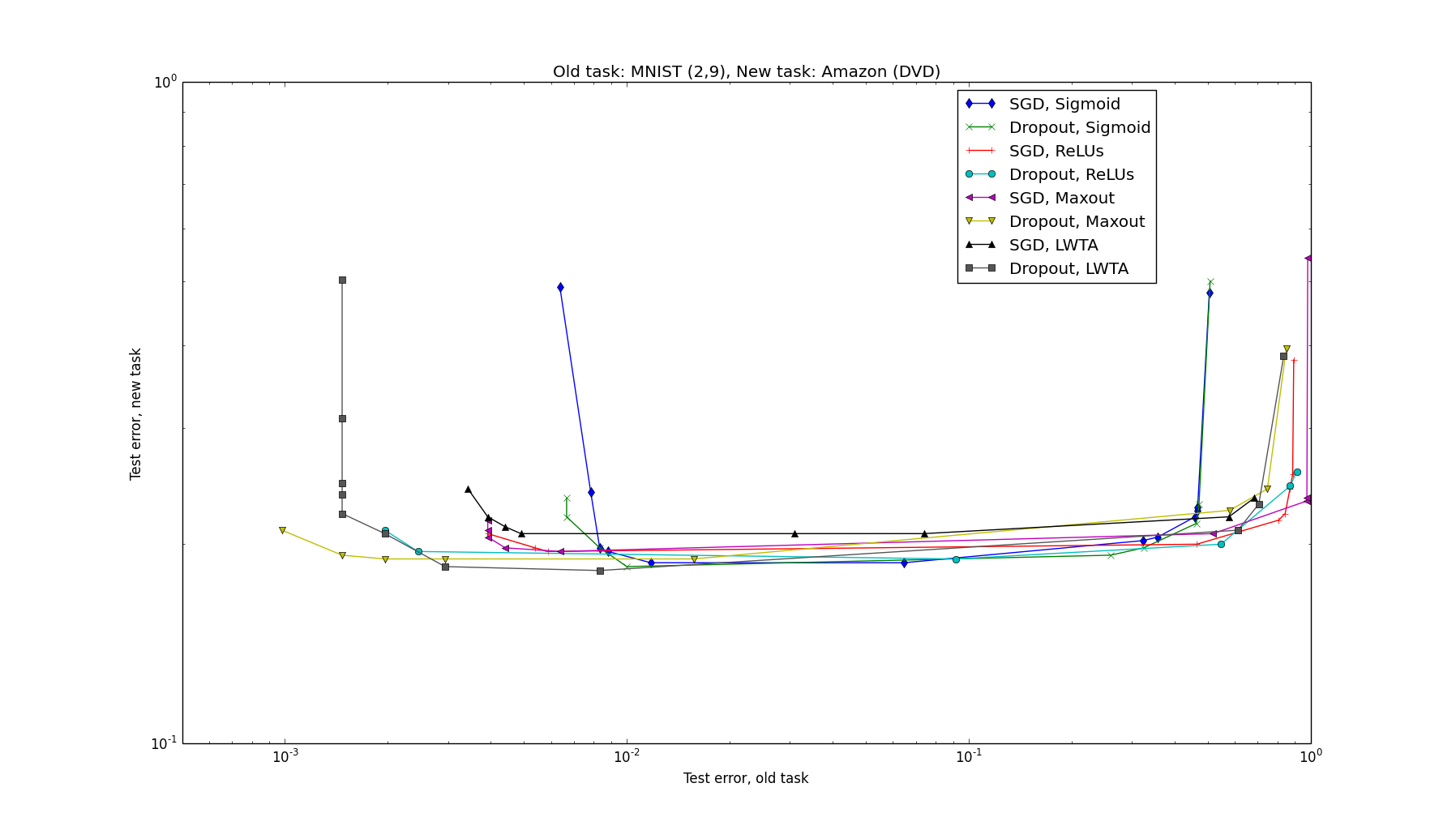}
\caption{Possibilities frontiers for the dissimilar tasks experiment.}
\label{dissimilar_tasks}
\end{figure*}

Classification performance results are presented
in Fig.~\ref{dissimilar_tasks}.
Using dropout improved the two-task validation set performance for all models on this task pair.
We show the effect of dropout on the optimal model size in Fig.~\ref{similar_size}.

\begin{figure}
\centering
\includegraphics[width=.4\textwidth]{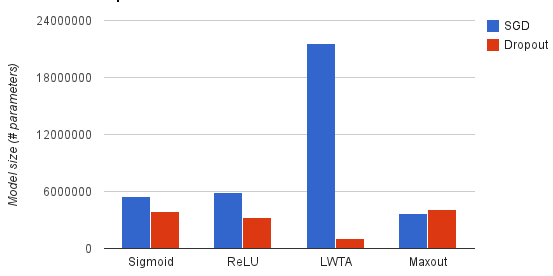}
\caption{Optimal model size with and without dropout on the disimilar tasks experiment.}
\label{similar_size}
\end{figure}

\section{Discussion}

Our experiments have shown that training with dropout is always beneficial, at least on the relatively
small datasets we used in this paper. Dropout improved performance for all eight methods on all three
task pairs. Dropout works the best in terms of performance on the new task, performance on the old task,
and points along the tradeoff curve balancing these two extremes, for all three task pairs.
Dropout's resistance to forgetting may be explained in part by the large model sizes that
can be trained with dropout. On the input-reformatted task pair and the similar task pair, dropout never decreased
the size of the optimal model for any of the four activation functions we tried. However, dropout seems to have
additional properties that can help prevent forgetting that we do not yet have an explanation for. On the
dissimilar tasks experiment, dropout improved performance but reduced the size of the optimal model for
most of the activation functions, and on the other task pairs, it occasionally had no effect on the optimal model size.

The only recent previous work on catastrophic forgetting\citep{Srivastava13} argued that the choice of activation function has a significant
effect on the catastrophic forgetting properties of a net, and in particular that hard LWTA outperforms logistic sigmoid
and rectified linear units in this respect when trained with stochastic gradient descent.

In our more extensive experiments we found that the choice of activation function
has a less consistent effect than the choice of training algorithm. When we performed experiments with different kinds of task
pairs, we found that the ranking of the activation functions is very problem dependent. For example, logistic
sigmoid is the worst under some conditions but the best under other conditions. This suggests that one should always
cross-validate the choice of activation function, as long as it is computationally feasible.
We also reject
the idea that hard LWTA is particular resistant to catastrophic forgetting in general, or that it makes the standard SGD training
algorithm more resistant to catastrophic forgetting. For example, when training with SGD on the input reformatting task pair,
hard LWTA's possibilities frontier is worse than all activation functions except sigmoid for most points along the curve. On the
similar task pair, LWTA with SGD is the worst of all eight methods we considered, in terms of best performance on the new task, best
performance on the old task, and in terms of attaining points close to the origin of the possibilities frontier plot.
However, hard LWTA does perform the best in some circumstances (it has the best performance on the new task for the dissimilar task pair
). This suggests that it is worth including hard LWTA as one of many activation functions in a hyperparameter search. LWTA is however
never the leftmost point in any of our three task pairs, so it is probably only useful in sequential task settings where forgetting is
an issue.

When computational resources
are too limited to experiment with multiple activation functions, we recommend using the maxout activation function trained with dropout. This is the only method that appears
on the lower-left frontier of the performance tradeoff plots for all three task pairs we considered.

\subsubsection*{Acknowledgments}
We would like to thank the developers of Theano~\citep{bergstra+al:2010-scipy,Bastien-Theano-2012},
Pylearn2~\citep{pylearn2_arxiv_2013}. We would also like to thank NSERC, Compute Canada, and Calcul Qu\'ebec
for providing computational resources. Ian Goodfellow is supported by the 2013 Google Fellowship in Deep
Learning.

\bibliography{strings,strings-shorter,ml,aigaion}
\bibliographystyle{icml2013}

\end{document}